\documentclass{article} 
\usepackage{nips12submit_e,times}
\usepackage{amsmath}
\usepackage{bbm}
\usepackage{graphicx}

\title{Recurrent Online Clustering as a Spatio-Temporal Feature Extractor in DeSTIN}

\author{
Steven R.~Young and Itamar~Arel
\\
Department of Electrical Engineering and Computer Science\\
University of Tennessee\\
\texttt{\{syoung22,itamar\}@eecs.utk.edu}
}

\nipsfinalcopy

\begin{document}

\maketitle

\begin{abstract}

This paper presents a basic enhancement to the DeSTIN deep learning architecture by replacing the explicitly calculated transition tables that are used to capture temporal features with a simpler, more scalable mechanism. This mechanism uses feedback of state information to cluster over a space comprised of both the spatial input and the current state. The resulting architecture achieves state-of-the-art results on the MNIST classification benchmark.
\end{abstract}

\section{Introduction}

We introduce an enhancement to DeSTIN \cite{KarnowskiPHD} that is aimed at simplifying the architecture and improving its ability to capture temporal features. The DeSTIN architecture consists of multiple instantiations of a common node arranged into layers. Using the recurrent clustering algorithm presented, spatiotemporal dependencies are naturally captured throughout the hierarchy. This recurrent clustering algorithm replaces the earlier incremental clustering algorithm and state-transition table which previously constructed the core of the DeSTIN node.  Each node operates independently and in parallel to all others which makes this architecture particularly well-suited to implementation in parallel hardware. The changes made here make the task of a parallel implementation, whether it be on a GPU or in custom analog circuitry, much more achievable.

\section{Recurrent Clustering Algorithm} 

In this section we will describe a recurrent incremental clustering
algorithm that forms centroids on a space comprised of a concatenation 
of the current spatial input and the previous belief state. The belief
state represents the probability that the current combination of external
observation and internal belief state belongs to each of the centroids.

\subsection{Incremental Clustering} 

The core of this recurrent clustering system is a winner-take-all
clustering algorithm that maintains estimates for the mean, $\mu$, and a variance,
$\sigma^{2}$, for each dimension of each centroid \cite{ITNG_2009}. These are updated
according to eqs. \eqref{eq:meanUpdate-1} and \eqref{eq:varUpdate-1}.
In these equations $o$ is the current observation, $x$ is the centroid
being updated, and $0<\alpha,\beta<1$ are learning rates. In order to
handle cases of poor centroid initialization, a mechanism called starvation
trace is employed to help select the centroid to be updated, as outlined in eqs. 
\eqref{eq:SelectUpdate-1} and \eqref{eq:starvTrace-1}. The starvation trace, $\psi$, 
decays with time when a centroid is not selected for an update. This value is used to weight
the distance between the corresponding centroid and the observation, such that centroids
which are distant from all observations can be updated.

\begin{equation}
\mu_{x}=\alpha\mu_{x}+(1-\alpha)(o-\mu_{x})\label{eq:meanUpdate-1}
\end{equation}
\begin{equation}
\sigma_{x}^{2}=\beta\sigma_{x}^{2}+(1-\beta)\left|(o-\mu_{x})^{2}-\sigma_{x}^{2}\right|\label{eq:varUpdate-1}
\end{equation}
\begin{equation}
x=\mbox{argmin\ensuremath{_{c\in C}\left[\psi_{c}\left\Vert o-\mu_{c}\right\Vert \right]}}\label{eq:SelectUpdate-1}
\end{equation}
\begin{equation}
\psi_{c}=\gamma\phi_{c}+(1-\gamma)\mathbbm{1}_{x=c}\label{eq:starvTrace-1}
\end{equation}

\subsection{Belief State Formulation} 

Once the selected centroid is updated, the elements of the belief state, $b_{c}$, are updated
using the normalized Euclidean distance between the $d$-dimensional input vector, $o$,
and each centroid, $c$, in the set of centroids, $C$, as shown in eqs. \eqref{eq:normDist} and \eqref{eq:belief}.

\begin{equation}
n_{c}=\sum_{i=1}^{d}\frac{(o_{i}-\mu_{c,i})^{2}}{\sigma_{c,i}^{2}}\label{eq:normDist}
\end{equation}
\begin{equation}
b_{c}=\frac{n_{c}^{-1}}{\sum\limits _{c^{\prime}\in C}n_{c'}^{-1}}\label{eq:belief}
\end{equation}


\subsection{Feedback Construct} 

\begin{figure}
\begin{centering}
\includegraphics[height=1.5 in]{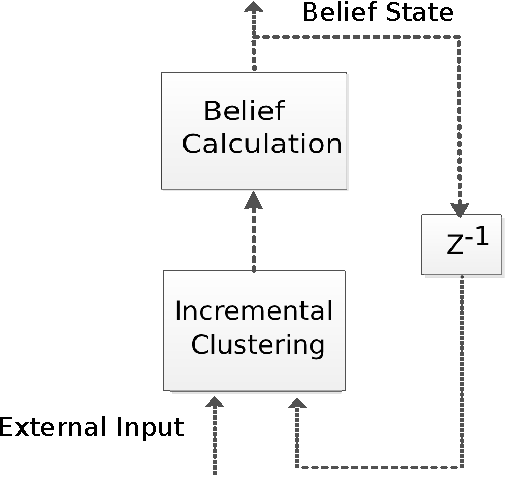}
\par\end{centering}

\caption{\label{fig:Recurrent-Clustering}In recurrent clustering the belief
state is concatenated with the current input to form the space over which clustering is performed.}
\end{figure}

This belief state is used as a portion of the input to the core
clustering algorithm, as depicted in Figure \ref{fig:Recurrent-Clustering}.
Introducing this feedback raises the concern of balancing the importance
of the spatial input and the temporal belief in the selection algorithm.
If the temporal component is not given adequate weight, this feedback loop
will have no effect. If, however, it is given too much weight, the spatial element
may be ignored. It was sufficient to employ 
a simple Euclidean distance measure to select the winning centroid. However, if the 
distributions of the spatial input and the belief states are vastly different,
it may be necessary to use a normalized Euclidean distance with a constant normalization
vector, the elements of which serve as weights for each dimension.

\section{Simulation Results}

Here we report on a series of experiments designed to demonstrate
the ability of the revised DeSTIN architecture to represent temporal
information. First, we explore the recurrent clustering algorithm's
abilities on a simple sequence analysis test case, and then the performance of the algorithm
in a full-scale architecture is demonstrated on a couple of standard benchmarks.

\subsection{Recurrent Clustering for Sequence Detection} 

\begin{figure}
\begin{centering}
\includegraphics[height=1.5 in]{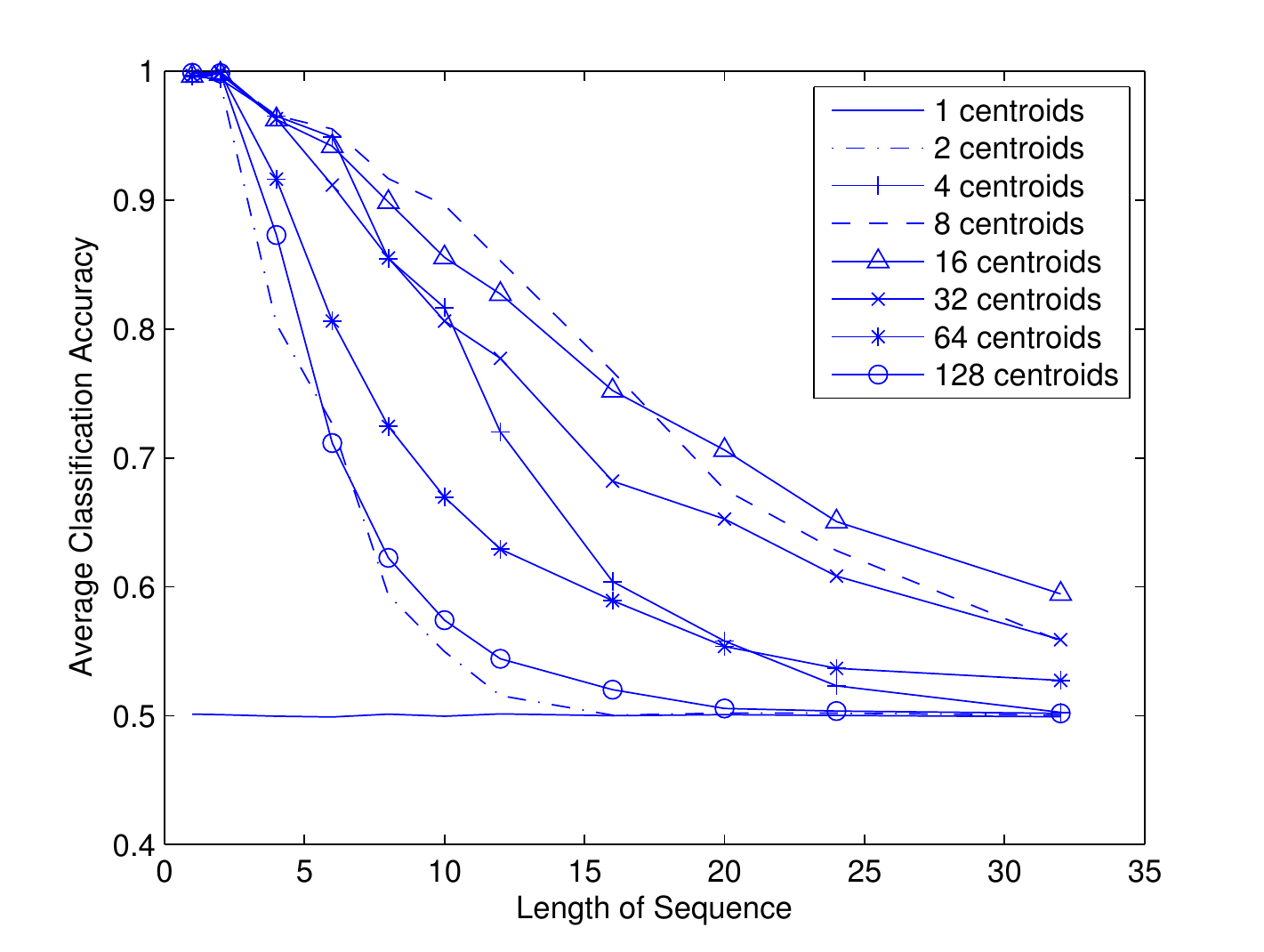}
\par\end{centering}

\caption{\label{fig:seqDet}Binary Sequence Detection: This plot shows average
classification accuracy for varying lengths of sequences.}
\end{figure}

It is important that the recurrent clustering be able to
capture regularities across time. In order to demonstrate this capability,
binary sequence detection tasks were studied with varying lengths
for the sequence of interest. The clustering algorithm is presented with equal probability either
the sequence of interest, or a sequence identical to it with only the first element
inverted. The belief states for each of the sequences were provided to a feed-forward neural
network for the purpose of classifying each sequence. Figure \ref{fig:seqDet} illustrates an exponentially-decaying relationship between
the classification accuracy and the length of the sequence of interest. This is to 
be expected given that there is no supervision to guide the algorithm in capturing temporal
dependencies of any specific length. The results highlight that if too few centroids are
considered, the algorithm has insufficient resources to represent dependencies that span longer time intervals. If too many centroids are used, the belief state may capture features in the data not relevant to identifying the sequence of interest. However,
such sensitivity to the number of centroids is desirably weak.

\subsection{MNIST Dataset}

The revised clustering scheme was next used in a full-scale architecture
and applied to the MNIST dataset classification problem \cite{mnistlecun}. 
The DeSTIN hierarchy used was made up of $3$ layers of nodes with each layer
consisting $4x4$, $2x2$ and $1$ nodes from bottom to top. The bottom layer
viewed a $16x16$ window that is shifted over the image, with each bottom layer
node receiving a unique $4x4$ patch. The bottom, middle, and top layer nodes used
$32$, $24$, and $32$ centroids, respectively. The hierarchy was then trained on $15,000$
of the training set images. Next, all training images and testing images were
provided to the hierarchy in order to generate feature vectors. The feature vector
for each image consisted of each node's belief state sampled at every $12th$ movement.
These feature vectors were provided to an ensemble of 11 feed-forward neural networks
trained with negative correlation learning. Each FFNN consisted of two hidden layers,
with 128 and 64 hidden neurons in the first and second layer respectively. A classification accuracy of $98.71\%$ 
was achieved which
is comparable to results using the first-generation DeSTIN architecture \cite{KarnowskiPHD} and
to results achieved with other state-of-the-art methods \cite{Kegl:2009:BPB:1553374.1553439,SalHinton07,ElasticDistortions}.



\bibliographystyle{ieeetr}
\addcontentsline{toc}{section}{\refname}{\footnotesize\bibliography{mil_bib}}
\end{document}